\documentclass{article}
\usepackage{amsmath,epsfig}
\usepackage[preprint]{spconfa4}
\usepackage{xcolor}
\usepackage{cite}

\usepackage{fancyhdr}

\let\OLDthebibliography\thebibliography
\renewcommand\thebibliography[1]{
  \OLDthebibliography{#1}
  \setlength{\parskip}{0pt}
  \setlength{\itemsep}{0pt plus 0.3ex}
}

\usepackage{amsfonts}
\usepackage{array}
\usepackage{amsthm,amsmath}
\usepackage{mathrsfs}
\usepackage{multirow}
\usepackage{url}
\usepackage{xspace}

\newcommand{\etal}{\emph{et al.}\xspace}
\begin{document}\sloppy

\def\x{{\mathbf x}}
\def\L{{\cal L}}

\title{GR-GAN: Gradual Refinement Text-to-image Generation}
%
\name{Bo Yang, Fangxiang Feng, Xiaojie Wang$^{\ast}$}
\address{Beijing University of Posts and Telecommunications, China\\
\{y\_b, fxfeng, xjwang\}@bupt.edu.cn}

\maketitle

\begin{abstract}
A good Text-to-Image model should not only generate high quality images, but also ensure the consistency between the text and the generated image. Previous models failed to simultaneously fix both sides well. This paper proposes a Gradual Refinement Generative Adversarial Network (GR-GAN) to alleviates the problem efficiently. A GRG module is designed to generate images from low resolution to high resolution with the corresponding text constraints from coarse granularity (sentence) to fine granularity (word) stage by stage, a ITM module is designed to provide image-text matching losses at both sentence-image level and word-region level for corresponding stages. We also introduce a new metric Cross-Model Distance (CMD) for simultaneously evaluating image quality and image-text consistency. Experimental results show GR-GAN significant outperform previous models, and achieve new state-of-the-art on both FID and CMD. A detailed analysis demonstrates the efficiency of different generation stages in GR-GAN. 
\end{abstract}
\section{INTRODUCTION}
\label{sec:intro}

\renewcommand{\thefootnote}{\fnsymbol{footnote}} 
\footnotetext[1]{Corresponding authors.} 
\footnotetext[2]{Code released at \url{https://github.com/BoO-18/GR-GAN.git}}

Text-to-Image synthesis aims to automatically generate images conditioned on text descriptions, which is one of the most popular and challenging multi-modal task. The task requires the generator not only generates high-quality images, but also preserve the semantic consistency between the text and the generated image.  
Generative Adversarial Networks (GANs) \cite{reed2016generative} have shown promising results on text-to-image generation by using the sentence vector as a conditional information. Zhang \etal \cite{zhang2018stackgan++} proposes Stack-GAN++, which employed a multi-stage structure to improve image resolution stage by stage, and an unconditional loss besides a conditional loss at each stage. Xu \etal \cite{xu2018attngan} proposes Attn-GAN with a module DAMSM to strengthen the consistency constraint on the generator. These models have achieved great improvements on the task, but the performances are still not satisfied, especially on complex scenes. 

There are several important problems in previous models. (1) The current multi-stage networks generate images from low resolution to high resolution for better image quality without considering the language constraints in corresponding granularities. The correspongdings between images in different resolutions and text descriptions in different granularities are nomrally specialized, images with low resolution are always described in coarse granularity, images with high resolution can be better described in fine granularity. The problem should be addressed for both better image quality and image-text consistency. (2) None of the current evaluation metrics could simultaneously evaluate image quality and image-text consistency, which are not suitable enough for this task. Meanwhile, previous works \cite{hinz2019semantic,li2019object,tao2020df} have already observed that the evaluation metric IS~\cite{salimans2016improved} and R-precision~\cite{xu2018attngan} of some models have serious overfitting, which makes it impossible to accurately evaluate the Text-to-Image model.

To address the above problems, we propose a Gradual Refinement Generate Adversarial Network (GR-GAN) including a Gradual Refinement Generator (GRG) and a Image-Text Matcher (ITM). GRG is a multi-stage structure. As the increasing of the resolution of generated images, the text constraints are refined from sentences (coarse granularity) to words (fine granularity) stage by stage. The first discriminator uses only unconditional loss so that it focuses on image quality, and following two discriminators use conditional loss so that they focus more on text-image consistency. ITM computes image-text matching losses at both sentence-image level and word-region level. Those losses are used in corresponding stages of GRG, i.e. sentence-image level loss is used in second stage and word-region level loss is used in third stage. We also design a novel metric named CMD (Cross Model Distance) for evaluating image quality and image-text consistency simultaneously. Experimental results show GR-GAN significant outperform previous models, and achieve new state-of-the-art on both FID and CMD. A detailed analysis demonstrates the efficiency of different generation stages in GR-GAN.

The main contributions can be summarized as follows:
\begin{itemize}
  \item We propose a GR-GAN model for progressively synthesizing images conditioned on text descriptions, where GRG module generates images from low resolution to high resolution with the corresponding text constraints from coarse granularity to fine granularity stage by stage, ITM provides different level of image-text matching losses at each stage.
  \item We propose a new Text-to-Image measurement CMD which can simultaneously evaluate Image Quality and Image-Text Consistency. It is more suitable for evaluating Text-to-Image task.
  \item Experimental results on MS-COCO dataset show that the GR-GAN significantly outperforms previous models, and achieve new state-of-the-art on both FID and CMD. 
\end{itemize}

\section{Related Work}
\label{sec:related}
\textbf{Text-to-Image Synthesis. }GAN-INT-CLS~\cite{reed2016generative} first proposed the use of conditional GAN to solve text-to-image task, which became a standard paradigm for subsequent work. Zhang \etal ~\cite{zhang2018stackgan++} proposed a multi-stage network to improve the resolution of the generated image stage by stage, which has made significant progress compared with the single-stage model. In order to make better use of language information, Attn-GAN~\cite{xu2018attngan} and DM-GAN~\cite{zhu2019dm} respectively proposed the use of word granularity information based on Attention and Dynamic Memory. The word granularity information can be used to help the generation of image regions, which effectively strengthens the image quality. Others ~\cite{hinz2019semantic, li2019object, hinz2019generating} further expand a object generation stage on the basis of multi-stage network to explicitly generate different objects. Compared to these multi-stage models, our proposed GR-GAN has a clearer division of work at each stage of the network by using a gradual refinement generator(see Sec.\ref{sec:GRG}).

\textbf{Image-Text Consistency. }Attn-GAN proposed a consistency constraint module DAMSM, which is pre-trained by image-text matching tasks on MS-COCO~\cite{lin2014microsoft}. By adding an additional consistency loss in the training process, image-text consistency is significantly improved. Since DAMSM has poor consistency assessment ability, Mirror-GAN~\cite{qiao2019mirrorgan} adds a text generation model on DAMSM to strengthen the image-text consistency. CP-GAN~\cite{liang2020cpgan} used a yoloV3 model to replace DAMSM, and used an object-level consistency constraint. These works enhanced the image-text consistency by strengthening the consistency constraint module. Our GR-GAN builds an ITM (see Sec.\ref{sec:ITM}) module that can calculate the similarity between images and text more accurately.

\textbf{Evaluation Metrics. }Evaluating Text-to-Image models need to consider both image quality and image-text consistency. Current methods use IS~\cite{salimans2016improved} and FID~\cite{heusel2017gans} to evaluate image quality, and R-precision~\cite{xu2018attngan} to evaluate the image-text consistency. However, \cite{li2019object,tao2020df} turns out that IS always fails on COCO~\cite{lin2014microsoft}. Our work also indicates a clear overfitting of IS metric(see Sec.\ref{sec:exper}). Previous works \cite{hinz2019semantic, frolov2021adversarial,zhang2021cross} find that some report R-precision scores significantly higher than real images. Our work found that this is due to the use of DAMSM in both training and evaluation. Due to the huge shortcomings of the current metrics, we design a novel indicator CMD that can simultaneously evaluating image quality and image-text consistency.

\section{PROPOSED METHOD}

\graphicspath{{Image/}}

\begin{figure*}[t]
    \centering
    \includegraphics[width=0.9\textwidth]{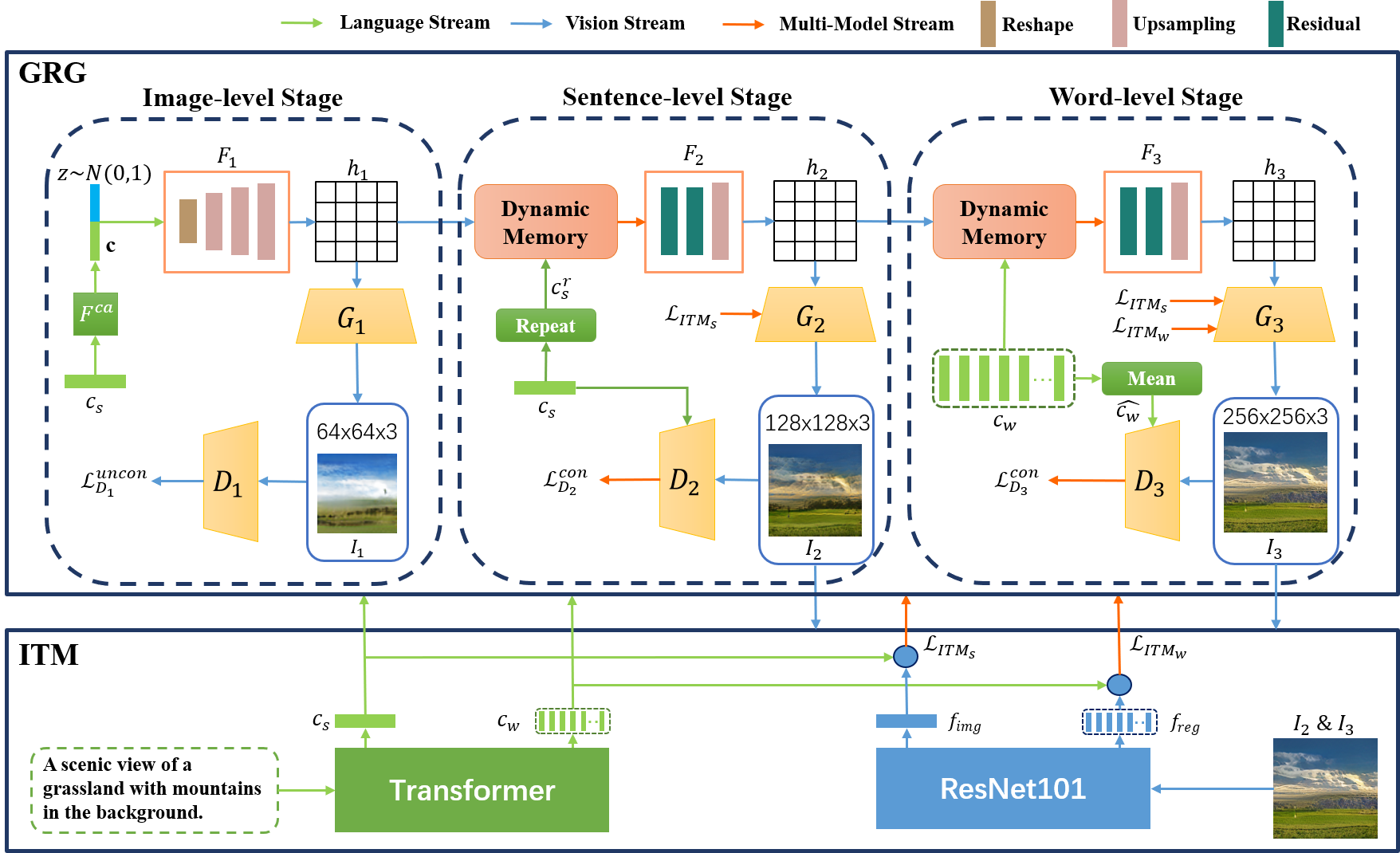}
    \caption{The overall structure of the GR-GAN model. The upper part is the GRG module, and the lower part is the ITM module.}
    \label{fig:Model}
\end{figure*}

The overall architecture of our GR-GAN model is shown in Fig.\ref{fig:Model}. The model mainly contains two core modules: ITM model and GRG model.

The GRG model includes three-level generators ($G_i, i=1,2,3$) and corresponding three-level discriminators ($D_i, i=1,2,3$). In all levels of networks, a neural network ($F_i, i=1,2,3$) is used to obtain a hidden image feature representation ($h_i, i=1,2,3$) which is as the input of the $G_i, i=1,2,3$ for generating corresponding images ($I_1, I_2, I_3$). It generates images with better quality as well as higher image-text consistency gradually.

The ITM model includes two parts. One part is for encoding of text descriptions and generated images, the other provides similarities (image-text matching losses) in different granularities between images and texts. 

\subsection{Gradual Refinement Generator}
\label{sec:GRG}
The GRG model consists of three stages:

\textbf{Image Initialize Stage: }An unconditional adversarial network ($D_1$,$G_1$) is employed for generating a initial image. The stage focuses more on quality of generated images. A $F^{ca}$ module is used to convert the sentence vector $c_s$ encoded by the transformer in ITM to the condition vector $c$. $F_1$ fuse $c$ with a random initial Gaussian noise $z$ to obtain a new image feature $h_1$, and $G_1$ use $h_1$ to generate a 64*64 image. $D_1$ is designed to judges whether the image is a generated image or a real image. The unconditional loss function is defined as follows:

\begin{footnotesize}
\begin{equation}
\begin{aligned}
    \mathcal{L}_{D_1}^{uncon}=&-\dfrac{1}{2} \mathbb{E}_{{x_1}\sim {data}}[log(D(x_1))]\\
    &-\dfrac{1}{2} \mathbb{E}_{{x_1}\sim {G_1}}[1 - log(D(x_1 ))]\\
    \mathcal{L}_{G_1}^{uncon} =& -\dfrac{1}{2} \mathbb{E}_{{x_1}\sim {G_1}}[log(D(x_1))]
\end{aligned}
\end{equation}
\end{footnotesize}


\textbf{Sentence-level Refinement Stage: } Sentence-level consistency constraints are introduced and enhanced in a conditional adversarial network ($D_2$,$G_2$) in this stage. The sentence encoding ${c_s} \in \mathcal{R}^{1\times N}$ is copied L times to ${c_s^r} \in \mathcal{R}^{L\times N}$(L is the max number of words set in code). And then a Dynamic Memory~\cite{zhu2019dm} is used to fuse ${c_s^r}$ with the image feature $h_1$, as shown in Equ.\ref{equ:DM1}:

\begin{footnotesize}
\begin{equation}
    T_{R-S} = F^{DM}(h_1, {c_s^r}) \label{equ:DM1}
\end{equation}
\end{footnotesize}

$T_{R-S}$ is then encoded by $F_2$ to obtain a new image feature $h_2$. $h_2$ is used by $G_2$ to generate a 128*128 image. At this stage, $D_2$ is designed to judge whether the input image matches the sentence description $c_s$. The conditional loss function is as Equ.\ref{equ:D2}:

\begin{footnotesize}
\begin{equation}
\begin{aligned}
    \mathcal{L}_{D_2}^{con}=&-\dfrac{1}{2} \mathbb{E}_{{x_2}\sim {data}}[log(D(x_2, c_s))]\\
    &-\dfrac{1}{2} \mathbb{E}_{{x_2}\sim {G_2}}[1 - log(D(x_2, c_s))] \label{equ:D2}\\
    \mathcal{L}_{G_2}^{con} =& -\dfrac{1}{2} \mathbb{E}_{{x_2}\sim {G_2}}[log(D(x_2, c_s))]
\end{aligned}
\end{equation}
\end{footnotesize}


We add another loss ${L}_{ITM_s}$ for sentence-image level consistency on ${G_2}$ in order to strengthen the constraints on the overall consistency between the image and the sentence at this stage. The loss is provided by ITM model and will be defined in details in Sec.\ref{sec:ITM}. The total sentence-image loss in this stage is therefore shown as in Equ.\ref{equ:G2}:

\begin{footnotesize}
\begin{equation}
    \mathcal{L}_{{G_2}-{total}}^{con} = \mathcal{L}_{G_2}^{con} + \lambda_2\mathcal{L}_{ITM_s} \label{equ:G2}
\end{equation}
\end{footnotesize}

\textbf{Word-level Refinement Stage:} Word-level consistency constraints are introduced and enhanced in a conditional adversarial network ($D_3$,$G_3$) in this stage to generate images with better image-text consistency in word-region level. Word embeddings $c_w$ encoded by the transformer in ITM model and $h_2$ generated in last stage are fused firstly by a Dynamic Memory~\cite{zhu2019dm} as shown in Equ.\ref{equ:DM2}: 

\begin{footnotesize}
\begin{equation}
    T_{R-W} = F^{DM}(h_2, {c_w}) \label{equ:DM2}
\end{equation}
\end{footnotesize}

A forward Neural network $F_3$ is then employed to map $T_{R-W}$ to a new representation $h_3$ as the input of $G_3$. At this stage, $D_3$ is designed to judge whether the input image is consistent with text description at the word level vector $\overline{c_w} \in \mathcal{R}^{1\times N}$ which is the average pooling of all word vectors $c_w$. The conditional loss function at this stage is as follows:

\begin{footnotesize}
\begin{equation}
\begin{aligned}
    \mathcal{L}_{D_3}^{con}=&-\dfrac{1}{2} \mathbb{E}_{{x_3}\sim {data}}[log(D(x_3, \overline{c_w}))]\\
    &-\dfrac{1}{2} \mathbb{E}_{{x_3}\sim {G_3}}[1 - log(D(x_3, \overline{c_w}))]\\
     \mathcal{L}_{G_3}^{con} =& -\dfrac{1}{2} \mathbb{E}_{{x_3}\sim {G_3}}[log(D(x_3, \overline{c_w}))]
\end{aligned}
\end{equation}
\end{footnotesize}


We also add consistency loss on ${G_3}$. Besides sentence-image level loss as in second stage, a word-region level loss ${L}_{ITM_w}$ is added to further strengthen the constraints on the fine granularity consistency between the image and the description. ${L}_{ITM_w}$ is also provided by ITM model and will be given in Sec.\ref{sec:ITM}. The total text-image loss in this stage is therefore shown as in Equ.\ref{equ:G3}:

\begin{footnotesize}
\begin{equation}
    \mathcal{L}_{{G_3}-{total}}^{con} = \mathcal{L}_{G_3}^{con} + \lambda_1\mathcal{L}_{ITM_s}+ \lambda_2\mathcal{L}_{ITM_w} \label{equ:G3}
\end{equation}
\end{footnotesize}

\begin{figure}[t]
    \centering
    \includegraphics[width=0.5\textwidth]{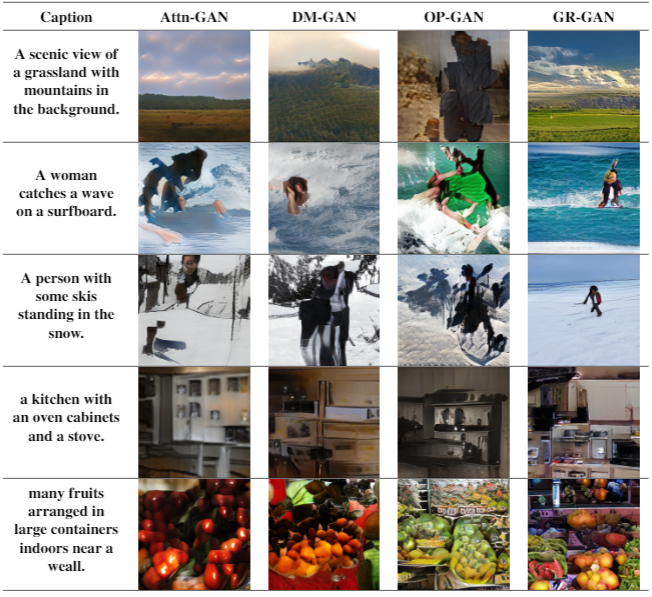}
    \caption{Generated images for selected examples from MS-COCO. GR-GAN generated images are generally of much higher quality and depict clearer scenes.}
    \label{fig:samples}
\end{figure}

\subsection{Image-Text Matcher}
\label{sec:ITM}

The ITM model is a consistency constraint model constructed by a Transformer~\cite{vaswani2017attention}and a ResNet101~\cite{he2016deep}, which is trained on the basis of CLIP~\cite{radford2021learning} parameters. Different from CLIP, ITM also extracts the image region features $f_{reg}$ and the word vectors $c_w$ together with the image feature $f_{img}$ and sentence vector $c_s$.

By using above information, ITM can provides sentence-image level similarity and word-region level similarity for images and text descriptions which are used as consistency losses in GRG model. The definitions of these losses are given as follows:

\begin{footnotesize}
\begin{equation}
    \mathcal{L}_1^s = - \sum_{i=1}^{M} \log \frac{\gamma exp(R(I_i,S_i))}{\sum_{i=1}^{M}\gamma exp(R(I_i,S_j))} \label{equ:LS}
\end{equation}
\end{footnotesize}

Where $R(I_i,S_j)$ is the matching score for image-sentence pair $(I_i,S_j)$. See appendix for calculation details. $\mathcal{L}_2^s$ can be obtained by replacing the denominator of the Equ.\ref{equ:LS} with ${\sum_{i=1}^{M}\gamma exp(R(I_j,S_i))}$. We can obtain $\mathcal{L}_1^w$ and $\mathcal{L}_2^w$ by calculating matching score for word-region pairs. The total ITM loss is as follow:

\begin{footnotesize}
\begin{equation}
\begin{aligned}
    \mathcal{L}_{ITM}&=\lambda_1\mathcal{L}_{ITM_s} + \lambda_2\mathcal{L}_{ITM_w}\\
    &=\lambda_1(\mathcal{L}_1^s + \mathcal{L}_2^s)+\lambda_2(\mathcal{L}_1^w + \mathcal{L}_2^w)
    \label{equ:ITMtotal}
\end{aligned}
\end{equation}
\end{footnotesize}

\subsection{Cross Model Distance}
\label{sec:CMD}
We propose Cross Model Distance(CMD) for simultaneously evaluating image quality and image-text consistency by mapping image and text information into a multi-modal semantic distribution. CMD is defined as follows:


\begin{footnotesize}
\begin{equation}
\begin{aligned}
    CMD = Dis(f,r) + \begin{vmatrix} Dis(f,l) - Dis(r,l) \end{vmatrix}\\
\end{aligned}
\end{equation}
\end{footnotesize}

Where, $f$ is the feature of the generated image, $r$ is the feature of the real image, $l$ is the feature of the text description. All Dis() are the $Fr\Acute{e}chet \  Distance$~\cite{frechet1957distance} between two distributions. Details can be see in appendix.



In the CMD, $\begin{vmatrix} Dis(f,l) - Dis(r,l) \end{vmatrix}$ measures the image-text consistency, denoted as \textbf{ITDis} for short, the smaller the \textbf{ITDis} is, the better the image-text consistency is. $Dis(f,r)$ measures the image quality, the smaller $Dis(f,r)$ is, the better the image quality is. 

\section{Experiments}

\begin{table}[t]
\centering
\begin{center}
\renewcommand{\arraystretch}{1.2}
\caption{Results on MS-COCO dataset ($\dag$ means the result is from the original paper)} \label{tab:results}
\begin{tabular}{ccccc}
  \hline
  & \textbf{IS$\uparrow$} & \textbf{FID$\downarrow$} & \textbf{ITDis$\downarrow$} & \textbf{CMD$\downarrow$} \\
  \hline
  Attn-GAN~\cite{xu2018attngan} & 23.61 & 34.90 & 1.07 & 15.97 \\
  \hline
  Obj-GAN$^\dag$~\cite{li2019object} & 24.07 & 36.52 & - & - \\
  \hline
  DM-GAN~\cite{zhu2019dm} & 32.32 & 27.42 & 0.98 & 12.68 \\
  \hline
  OP-GAN~\cite{hinz2019semantic} & 27.88 & 24.70 & 0.83	& 11.99 \\
  \hline
  CP-GAN~\cite{liang2020cpgan} & \textbf{52.73} & 47.91 & 1.27 & 16.05 \\
  \hline
  DALL-E$^\dag$~\cite{ramesh2021zero} & 17.90 & 27.50 & - & - \\
  \hline
  GR-GAN & 25.60 & \textbf{22.58} & \textbf{0.80} & \textbf{8.04} \\
  \hline
\end{tabular}
\end{center}
\end{table}

\subsection{Experimental settings}

Following previous work, we report validation results by generating images for 30,000 random captions on MS-COCO. 


\textbf{Evaluation metrics: }As previous work did, Inception Score (IS)~\cite{salimans2016improved} and $Fr\Acute{e}chet\ Inception\ Distance$ (FID)~\cite{heusel2017gans} are used to evaluate image quality. Our work further proves that since the calculation method of IS evaluation is relatively simple, it is not suitable for datasets with complex scenes such as MS-COCO. Since R-Precision~\cite{xu2018attngan} has huge drawbacks as Sec.\ref{sec:related} report, but ITDis defined in Sec.\ref{sec:CMD} is direct and intuitive, we use ITDis for image-text consistency evaluation. CMD defined in Sec.\ref{sec:CMD} is used for evaluating image quality and image-text consistency as a whole. 
To ensure the fairness of evaluation, CMD and ITDis use a off-the-shelf CLIP-ViT network for evaluation, which is different from our CLIP-Res101 used in ITM.

\textbf{Parameter settings: } Our work refers to Xu \etal \cite{xu2018attngan} and set $\gamma$ in Equ.\ref{equ:LS} to 10. For $\lambda_1$ and $\lambda_2$ in Equ.\ref{equ:ITMtotal}, we set them between 1 and 5 for parameter adjustment experiments. Finally, it is determined that $\lambda_1$ is 4 and $\lambda_2$ is 1. The model learning rate and training epochs are set to 0.0002 and 300.

\subsection{Experimental results}
\label{sec:exper}

Table.\ref{tab:results} gives the experimental results of our GR-GAN model and multiple current mainstream models on MS-COCO dataset. It's easily to see that CP-GAN could get high IS but perform poorly on the stronger FID, which indicates a clear overfitting of IS because of ignoring the real image distribution. Our model achieves best performance on three metrics. The FID metric of the GR-GAN model is increased by 8.58$\%$ (24.70 to 22.58) compared with the previous best OP-GAN model, increased by 17.65$\%$ (27.42 to 22.58) compared with DM-GAN. At the same time, GR-GAN is superior to the existing models in the consistency metric ITDis, and has a significant improvement in the comprehensive metric CMD compared with other models (11.99 to 8.04). These experimental results show that our GR-GAN is more effective than the previous SOTA model to generate high-quality images as well as with high image-text consistency. 

Fig.\ref{fig:samples} gives some examples. As we can see, both the image quality and image-text consistency of our model are significantly better than those in previous models.

\begin{table}[t]
\centering
\begin{center}
\caption{Results of ablation studies. (DA means using DAMSM as the consistency constraint module, IT means using ITM. GRG refers to adding the GRG method)} \label{tab:ablation}
\renewcommand{\arraystretch}{1.0}
\begin{tabular}{ccccccc}
  \hline
  \textbf{backbone} & \textbf{DA} & \textbf{IT} & \textbf{GRG} & \textbf{FID} & \textbf{ITDis} & \textbf{CMD}\\
  \hline
     \multirow{5}*{Attn-GAN}
     & & & & 43.88 & 1.31 & 17.65 \\
     \cline{2-7}
     & \checkmark & & & 34.90 & 1.07 & 15.97 \\ 
     \cline{2-7}
     & \checkmark & & \checkmark & 29.69 & 1.05 & 14.89 \\
     \cline{2-7}
     & & \checkmark & & 30.21 & 1.04 & 11.36 \\
     \cline{2-7}
     & & \checkmark & \checkmark & 28.72 & 0.87 & 10.57 \\
  \hline
     \multirow{5}*{DM-GAN}
     & & & & 28.56 & 1.06 & 13.67 \\
     \cline{2-7}
     & \checkmark & & & 27.42 & 0.98 & 12.68 \\ 
     \cline{2-7}
     & \checkmark & & \checkmark & 25.83 & 1.00 & 11.84 \\
     \cline{2-7}
     & & \checkmark & & 25.33 & 0.96 & 9.01 \\
     \cline{2-7}
     & & \checkmark & \checkmark & 22.58 & 0.80 & 8.04 \\
  \hline
\end{tabular}
\end{center}
\end{table}

\subsection{Ablation studies}

We conducted some ablation experiments on the two core modules (GRG and ITM) on Attn-GAN and DM-GAN frame respectively.

\begin{figure}[t]
    \centering
    \includegraphics[width=0.5\textwidth]{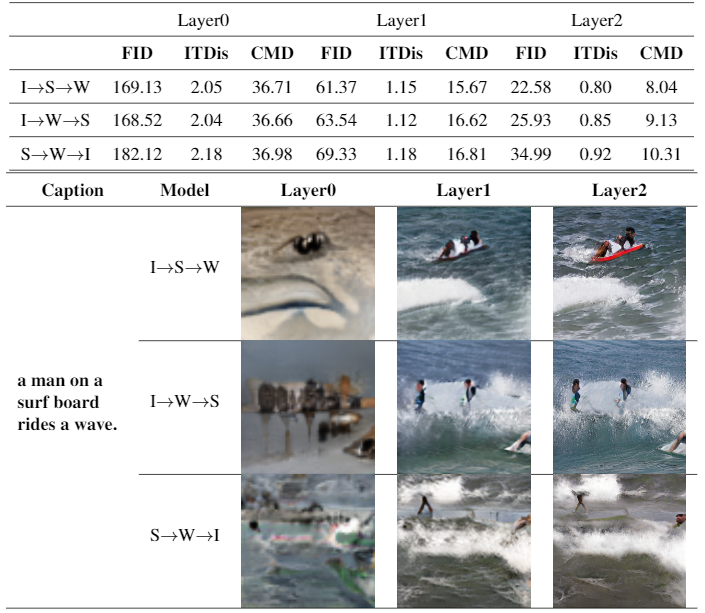}
    \caption{Performance under different order ($I$: Image Stage. $S$: Sentence-level Stage. $S$: Word-level Stage.)}
    \label{fig:ablation}
\end{figure}

\textbf{GRG: }As shown in Table.\ref{tab:ablation}, it can be seen that adding GRG not only effectively improve image quality (FID), but also generates images of better image-text consistency (ITDis), and achieves better performances on CMD.  
At the same time, we check the order of sentence and word constraints in the network, the experimental results and some examples are shown in Fig.\ref{fig:ablation}. As the first row shows, by using our gradual sequence, the generation process has better performance. First, the overall image distribution is generated, then the sentence-level global semantics are enriched, and the detailed information (such as surf board, wave) is perfected finally. When the order of the constraints is changed, both the quality and consistency are worse. It shows that our coarse-to-fine order to refine the image gradually is important to the model. In addition, the effect of putting the $I$ stage for image quality after the semantic stage $(S,W)$ is much more worse than putting it before the semantic layer (the third rows). It shows that for the low-resolution generation stage, it is more reasonable to focus on quality of images. 

\textbf{ITM: }In order to evaluate the role of our ITM module, we use ITM to replace DAMSM on Attn-GAN and DM-GAN respectively (the first and third rows of data in Table.\ref{tab:ablation}). The experimental results show that ITM brings significant improvement on image quality metric FID, as well as the comprehensive metric CMD. 

Ablation results show that both GRG and ITM can significantly improve the performance of FID and CMD, and using them together could get the best results.  

\section{Conclusions}
In this paper, we propose an Gradual Refinement Generative Adversarial Network (GR-GAN) for text-to-image synthesis. It generates images conditional on text descriptions gradually in three stages. Firstly, We build a novel GRG module to improve image quality by using language information from coarse-grained to fine-grained. Secondly, we propose a ITM model to compute image-text matching loss for training the generators of the GR-GAN, which is able to more accurately assess the image-text consistency than DAMSM. Finally, we propose an evaluation metric that is more suitable for Text-to-Image, which can simultaneously evaluate image quality and image-text consistency. Our GR-GAN significantly outperforms previous SOTA GAN models, boosting the best reported FID from 24.70 to 22.58 and CMD from 11.99 to 8.04 on MS-COCO. Extensive experimental results clearly demonstrate the effectiveness of our proposed methods.

\section{Acknowledgments}
This work was supported by the National Key Research and Development Program of China under Grant 2020YFF0305302, NSFC (No.61906018, No.62076032), the Central Universities under Grant 2021RC36 and the Cooperation Project with Beijing SanKuai Technology Co.,Ltd.

\bibliographystyle{IEEEbib}
\bibliography{ref}

\def\x{{\mathbf x}}
\def\L{{\cal L}}

\section{A Supplementary material}

In this section, we provide additional implementation details about our methods.

\subsection{Detailed Methods}

\textbf{ITM: } We mentioned that our ITM was built on the basis of CLIP parameters. CLIP offered four kinds of backbone to use,  but none of them could get the word embedding and the image region feature, which means that it's unable to calculate a fine granularity consistent loss with a original CLIP. For this reason, we use the parameters of the original CLIP-Res101, and add new neural network layers to calculate our region features and word vectors by training a image-text matching task on MS-COCO. Specifically, we intercept the output features from layer3 as the initial regional features, and use a 1*1 convolutional layer to fine-tune the word-region level loss. On the language side, we use a MLP layer and a Layer Normalization layer to calculate the word vector based on the token vector output by the Transformer.

Since the ITM was fine-tuned on MS-COCO and added a word-region level consistency, ITM could get a much more better results on Image-Text Retrieval task than CLIP and DAMSM. Retrieval results could be seen in table \ref{tab:retrieval}

\begin{table}[h]
\centering
\begin{center}
\renewcommand{\arraystretch}{1.5}
\caption{Image-Text Retrieval results for different pre-trained models on MS-COCO.} \label{tab:retrieval}
\begin{tabular}{ccccccc}
  \hline
  & \multicolumn{3}{c}{Image Retrieval Text} & \multicolumn{3}{c}{Text Retrieval Image} \\
  \cline{2-7}
  & \textbf{R1} & \textbf{R5} & \textbf{R10} & \textbf{R1} & \textbf{R5} & \textbf{R10} \\
  \hline
  DAMSM & 27.2 & 63.7 & 79.6 & 28.92 & 63.98 & 77.88 \\
  \hline
  CLIP-VIT & 49.6 & 77.4 & 87.8	& 47.14	& 77.26	& 88.24 \\
  \hline
  CLIP-Res & 49.6 & 78.7 & 87.9 & 47.64	& 78.20 & 87.98 \\
  \hline
  ITM & \textbf{53.1} & \textbf{83.8} & \textbf{92.7} & \textbf{52.64} & \textbf{83.22} & \textbf{92.60} \\
  \hline
\end{tabular}
\end{center}
\end{table}

\begin{figure}[t]
    \centering
    \includegraphics[width=0.5\textwidth]{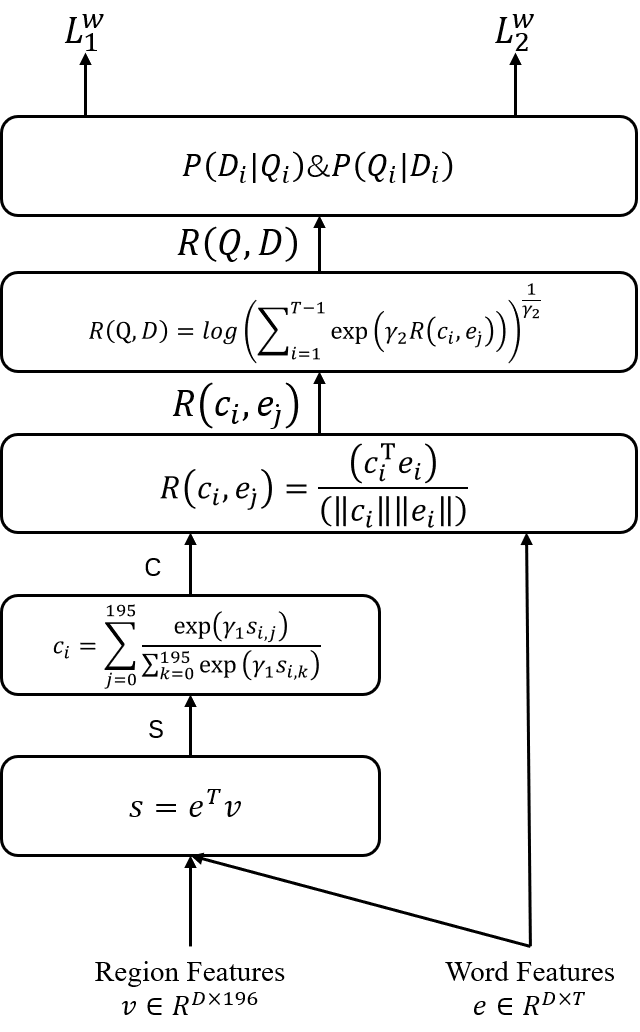}
    \caption{Detail of the Word-Region level loss calculate.}
    \label{fig:wordloss}
\end{figure}

The word granularity score calculation method is similar with DAMSM, and fig \ref{fig:wordloss} shows the detail of this method. We first use the image text feature to calculate the dot product and then use the attention method to calculate the interactive feature representation $c$ of a image-text area. We calculate the cosine distance between $c$ and the Region feature to characterize the degree of relevance of each word to the image area. Based on this similarity matrix, the final required image-text matching score R(Q, D) is calculated, where Q represents the generated image and D represents the text description. The loss function mainly focuses on the degree of matching between the generated image and the text description. For each image-text pair in the batch, we calculate the posterior probability of the match between the sentence description $D_i$ and the image $Q_i$.

\subsection{Cross Model Distance}

Our Cross Model Distance(CMD) is calculated based on a pre-trained CLIP-ViT. CMD map image and text information into a multi-modal semantic space, and simultaneously evaluate Image Quality and Image-Text Consistency by calculating the comprehensive feature space distance between the text, the generated image and the real image. 

CMD uses $Fr\Acute{e}chet\ Distance$ to calculate the distance between the real image and the generated image distribution, adding text distribution in the feature space, which is more in line with the Text-to-Image task. In the Text-to-Image task, we need to evaluate both the image quality and the image-text consistency. For this reason, we hope that the distribution distance between the generated image and the real image is as small as possible(the smaller the quality of generated image is better). At the same time, we hope that the difference between the distance between the generated image and the language distribution and the distance between the real image and the language distribution is as small as possible (the smaller the image and text consistency is better). From this, the calculation formula of CMD can be obtained as follows:

{\setlength\abovedisplayskip{0pt}
\setlength\belowdisplayskip{1pt}
\begin{footnotesize}
\begin{eqnarray}
\begin{aligned}
    &bias = \begin{vmatrix} Dis(f,l) - Dis(r,l) \end{vmatrix}\\
    &CMD = Dis(f,r) + bias\\
    &where\quad Dis(a,b) = \begin{Vmatrix}{\mu_a} - {\mu_b}\end{Vmatrix}^2 + Tr(\varepsilon_a + \varepsilon_b - 2\sqrt{\varepsilon_a\varepsilon_b})
\end{aligned}
\end{eqnarray}
\end{footnotesize}}

Here, $f$ represents the feature of the generated image, $r$ represents the feature of the real image, $l$ represents the feature of the text description, $\mu$ represents the mean value of the distribution, and $\varepsilon$ represents the covariance matrix of the distribution. The three distance calculations are based on the $Fr\Acute{e}chet\ Distance$. We hope that $Dis(f,r)$ is as small as possible(the smaller the Image Quality is better). At the same time, we hope that the difference between $Dis(f,l)$ and $Dis(r,l)$ is as small as possible (the smaller the Image-Text Consistency is better). The formula can be further derived to get the expression of CMD as:

\begin{eqnarray}
\begin{aligned}
    CMD = \left\{
    \begin{aligned}
        &2[(\mu_f - \mu_r)(\mu_f - \mu_l) + Tr(\varepsilon_f - \sqrt{\varepsilon_f\varepsilon_l}\\
        &+ \sqrt{\varepsilon_r\varepsilon_l} - \sqrt{\varepsilon_f\varepsilon_r})]\quad where\  bias > 0,
        \\
        &2[(\mu_r - \mu_f)(\mu_r - \mu_l) + Tr(\varepsilon_f + \sqrt{\varepsilon_f\varepsilon_l}\\
        &- \sqrt{\varepsilon_r\varepsilon_l} - \sqrt{\varepsilon_f\varepsilon_r})]\quad where\  bias < 0
    \end{aligned}
    \right.
\end{aligned}
\end{eqnarray}

When bias is zero, we have $\mu_l = ({\mu_r + \mu_f})/2$ and $\sqrt{\varepsilon_f\varepsilon_l} - \sqrt{\varepsilon_r\varepsilon_l} = ({\varepsilon_f - \varepsilon_r})/2$. In this case, the CMD index will degenerate into a CLIP-ViT-based FID index calculation. It is believed that the smaller the distance between the real image and the generated image, the better the quality of the generated image. When the generated image distribution is completely consistent with the real image distribution, we can get $\mu_r = \mu_f$ and $\varepsilon_f = \varepsilon_r$, so that the final calculation result of CMD is 0. It proves that our CMD index is a distance calculation method greater than or equal to 0. When the CMD is smaller, it indicates that the generated image distribution is closer to the real image distribution. 

In the CMD index, we regard $\begin{vmatrix} Dis(f,l) - Dis(r,l) \end{vmatrix}$ as the Image-Text Consistency index, denoted as \textbf{ITDis}, and regard Dis(f,r) as the index of Image Quality. By combining the two, we can obtain a \textbf{CMD} calculation scheme that can evaluate Image Quality and Image-Text Consistency.

\section{Additional experimental details}

In this chapter, we will show more generated examples and the clear results of ablation experiments. 

\begin{table}[h]
\centering
\begin{center}
\renewcommand{\arraystretch}{1.5}
\caption{Performance between the parameter experiments results of our GR-GAN on MS-COCO.(These experiments are tested by using ITM without GRG)} \label{tab:results}
\begin{tabular}{cccc}
  \hline
  \textbf{Setting}& \textbf{FID$\downarrow$} & \textbf{ITDis$\downarrow$} & \textbf{CMD$\downarrow$} \\
  \hline
  $\lambda_1 = 1,\ \lambda_2 = 1$ & 28.03 & 0.89 & 9.52 \\
  \hline
  $\lambda_1 = 2,\ \lambda_2 = 1$ & 26.58 & 0.91 & 9.55 \\
  \hline
  $\lambda_1 = 3,\ \lambda_2 = 1$ & 26.59 & 0.88 & 9.53 \\
  \hline
  $\lambda_1 = 4,\ \lambda_2 = 1$ & \textbf{25.33} & \textbf{0.87} & \textbf{8.93} \\
  \hline
  $\lambda_1 = 4,\ \lambda_2 = 2$ & 28.92 & 0.95 & 10.13 \\
  \hline
  $\lambda_1 = 5,\ \lambda_2 = 2$ & 29.92 & 0.97 & 10.02 \\
  \hline
\end{tabular}
\end{center}
\end{table}

\textbf{Generate Images: }Visual inspection of selected images in Fig \ref{tab:samples} show that the quality is greatly improved compared to previous models, and GR-GAN could generate much more realistic images than previous models.

\textbf{Parameter Experiments: }We conducted a lot of experiments to find suitable weight loss parameters for $\lambda_1$ and $\lambda_2$. Our experience is to set $\lambda_2$ to be smaller than $\lambda_1$, and the model always achieves the best results between 1 and 5. Table \ref{tab:results} shows the results of some parameter experiments, and we find that setting $\lambda_1$ to 4 and $\lambda_2$ to 1 can achieve the best results. At the same time, the results of the parameter experiment show that our CMD and ITDis metrics will not increase directly with the increase of loss weight, effectively avoiding the problem of R-Precision. It further illustrates the rationality of the metrics we proposed.

 \textbf{Adjust Sequence: }In order to demonstrate the rationality and effectiveness of our gradual refinement structure, we adjusted the order of the network in the experiment, and obtained the experimental results as shown in table \ref{tab:ablation}. As the first row shows, by using our gradual sequence, the generation process has better performance and progress. First, the overall image distribution is generated, then the overall semantics are enriched, and the detailed information (such as surf board, wave) is perfected at the end. Then we try to adjust the order of the semantic processing stages so that the model uses word information first, and then sentence information(the second row). Compared with the Gradual Structure we proposed, the optimization speed between stages becomes slower, and the final effect is a bit worse. It shows that when using semantic information to optimize an image, the semantic information should be used from coarse-grained to fine-grained to refine the image gradually. In addition, the effect of putting the $I$ stage for image quality after the semantic stage $(S,W)$ is much more worse than putting it before the semantic layer(the third rows). It shows that for the low-resolution generation stage, it is a more reasonable solution to focus on generating higher-quality images. 

\begin{table*}[t]
\centering
\begin{center}
\setlength{\abovecaptionskip}{0pt}%
\setlength{\belowcaptionskip}{10pt}%
\caption{Generated images for selected examples from MS-COCO. GR-GAN generated images are generally of much higher quality and depict clearer scenes. More random samples are available in the appendix.} \label{tab:samples}
\renewcommand{\arraystretch}{1.5}
\begin{tabular}{m{2.5cm}<{\centering}cccc}
  \hline
  \hline
  \textbf{Caption} & \textbf{Attn-GAN} & \textbf{DM-GAN} & \textbf{OP-GAN} & \textbf{GR-GAN}\\
  \hline
  \textbf{A scenic view of a grassland with mountains in the background.} 
    & 
    \begin{minipage}[b]{0.3\columnwidth}
		\centering
		\raisebox{-.5\height}{\includegraphics[width=\linewidth]{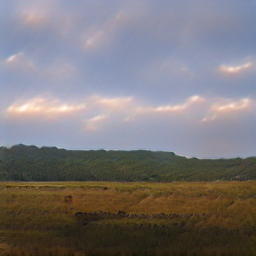}}
	\end{minipage} 
	&
	\begin{minipage}[b]{0.3\columnwidth}
		\centering
		\raisebox{-.5\height}{\includegraphics[width=\linewidth]{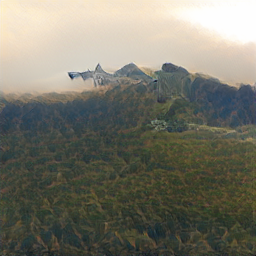}}
	\end{minipage} 
	&
	\begin{minipage}[b]{0.3\columnwidth}
		\centering
		\raisebox{-.5\height}{\includegraphics[width=\linewidth]{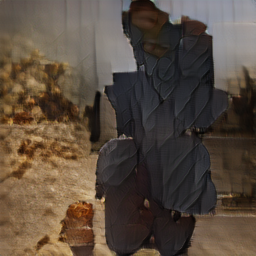}}
	\end{minipage} 
	&
	\begin{minipage}[b]{0.3\columnwidth}
		\centering
		\raisebox{-.5\height}{\includegraphics[width=\linewidth]{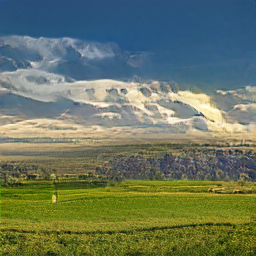}}
	\end{minipage} \\
  \hline
  
  \textbf{A woman catches a wave on a surfboard.} 
    & 
    \begin{minipage}[b]{0.3\columnwidth}
		\centering
		\raisebox{-.5\height}{\includegraphics[width=\linewidth]{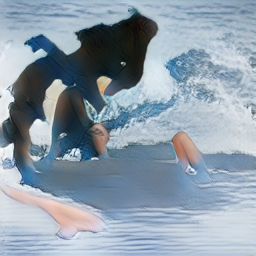}}
	\end{minipage} 
	&
	\begin{minipage}[b]{0.3\columnwidth}
		\centering
		\raisebox{-.5\height}{\includegraphics[width=\linewidth]{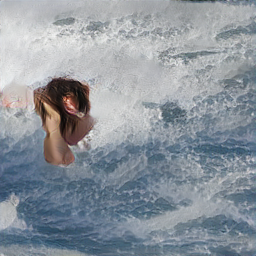}}
	\end{minipage} 
	&
	\begin{minipage}[b]{0.3\columnwidth}
		\centering
		\raisebox{-.5\height}{\includegraphics[width=\linewidth]{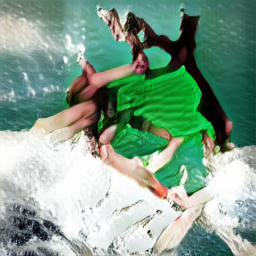}}
	\end{minipage} 
	&
	\begin{minipage}[b]{0.3\columnwidth}
		\centering
		\raisebox{-.5\height}{\includegraphics[width=\linewidth]{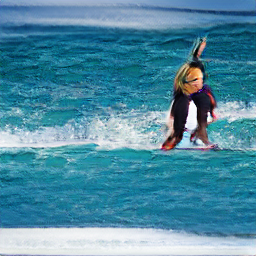}}
	\end{minipage} \\
  \hline
  
  \textbf{A person with some skis standing in the snow.} 
    & 
    \begin{minipage}[b]{0.3\columnwidth}
		\centering
		\raisebox{-.5\height}{\includegraphics[width=\linewidth]{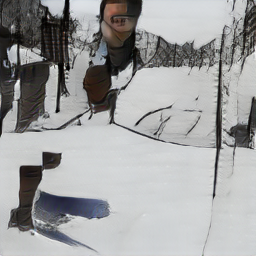}}
	\end{minipage} 
	&
	\begin{minipage}[b]{0.3\columnwidth}
		\centering
		\raisebox{-.5\height}{\includegraphics[width=\linewidth]{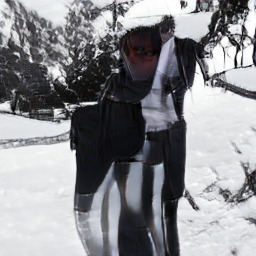}}
	\end{minipage} 
	&
	\begin{minipage}[b]{0.3\columnwidth}
		\centering
		\raisebox{-.5\height}{\includegraphics[width=\linewidth]{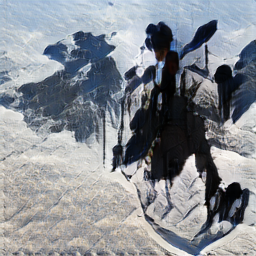}}
	\end{minipage} 
	&
	\begin{minipage}[b]{0.3\columnwidth}
		\centering
		\raisebox{-.5\height}{\includegraphics[width=\linewidth]{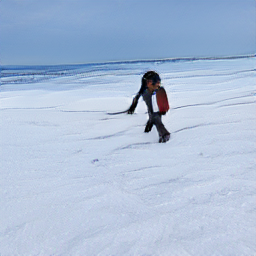}}
	\end{minipage} \\
  \hline
  
  \textbf{a kitchen with an oven cabinets and a stove.} 
    & 
    \begin{minipage}[b]{0.3\columnwidth}
		\centering
		\raisebox{-.5\height}{\includegraphics[width=\linewidth]{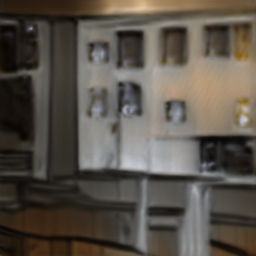}}
	\end{minipage} 
	&
	\begin{minipage}[b]{0.3\columnwidth}
		\centering
		\raisebox{-.5\height}{\includegraphics[width=\linewidth]{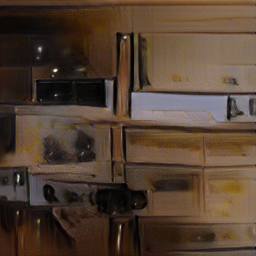}}
	\end{minipage} 
	&
	\begin{minipage}[b]{0.3\columnwidth}
		\centering
		\raisebox{-.5\height}{\includegraphics[width=\linewidth]{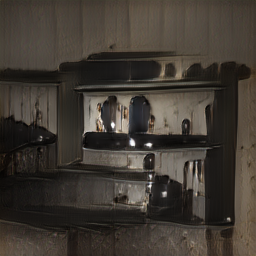}}
	\end{minipage} 
	&
	\begin{minipage}[b]{0.3\columnwidth}
		\centering
		\raisebox{-.5\height}{\includegraphics[width=\linewidth]{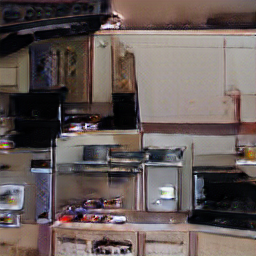}}
	\end{minipage} \\
  \hline
  
  \textbf{many fruits arranged in large containers indoors near a weall.} 
    & 
    \begin{minipage}[b]{0.3\columnwidth}
		\centering
		\raisebox{-.5\height}{\includegraphics[width=\linewidth]{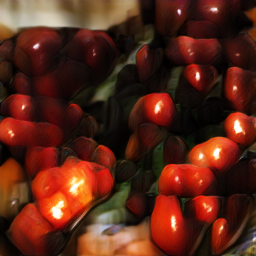}}
	\end{minipage} 
	&
	\begin{minipage}[b]{0.3\columnwidth}
		\centering
		\raisebox{-.5\height}{\includegraphics[width=\linewidth]{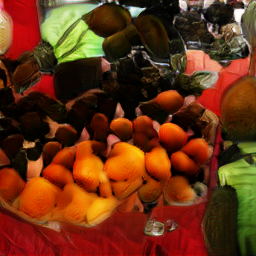}}
	\end{minipage} 
	&
	\begin{minipage}[b]{0.3\columnwidth}
		\centering
		\raisebox{-.5\height}{\includegraphics[width=\linewidth]{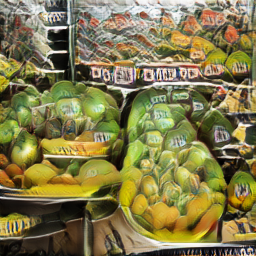}}
	\end{minipage} 
	&
	\begin{minipage}[b]{0.3\columnwidth}
		\centering
		\raisebox{-.5\height}{\includegraphics[width=\linewidth]{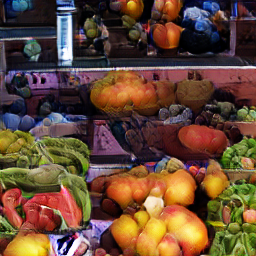}}
	\end{minipage} \\
  \hline
  
  \textbf{Many kites being flown in an open field.} 
    & 
    \begin{minipage}[b]{0.3\columnwidth}
		\centering
		\raisebox{-.5\height}{\includegraphics[width=\linewidth]{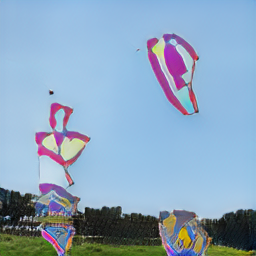}}
	\end{minipage} 
	&
	\begin{minipage}[b]{0.3\columnwidth}
		\centering
		\raisebox{-.5\height}{\includegraphics[width=\linewidth]{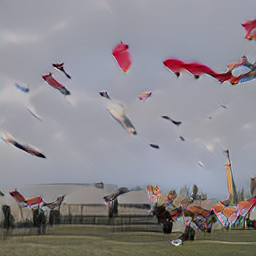}}
	\end{minipage} 
	&
	\begin{minipage}[b]{0.3\columnwidth}
		\centering
		\raisebox{-.5\height}{\includegraphics[width=\linewidth]{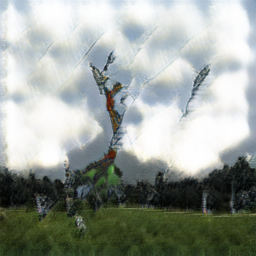}}
	\end{minipage} 
	&
	\begin{minipage}[b]{0.3\columnwidth}
		\centering
		\raisebox{-.5\height}{\includegraphics[width=\linewidth]{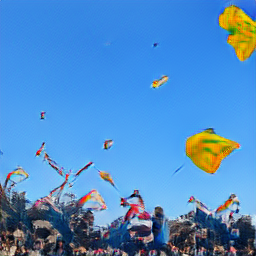}}
	\end{minipage} \\
  \hline
  
  \textbf{a busy city with cares and mopeds everywhere.} 
    & 
    \begin{minipage}[b]{0.3\columnwidth}
		\centering
		\raisebox{-.5\height}{\includegraphics[width=\linewidth]{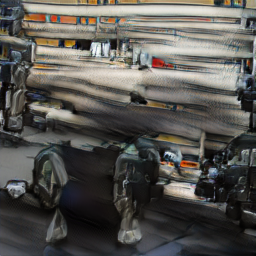}}
	\end{minipage} 
	&
	\begin{minipage}[b]{0.3\columnwidth}
		\centering
		\raisebox{-.5\height}{\includegraphics[width=\linewidth]{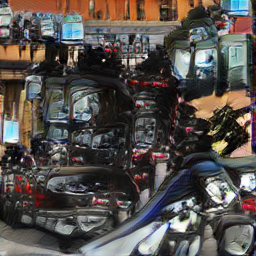}}
	\end{minipage} 
	&
	\begin{minipage}[b]{0.3\columnwidth}
		\centering
		\raisebox{-.5\height}{\includegraphics[width=\linewidth]{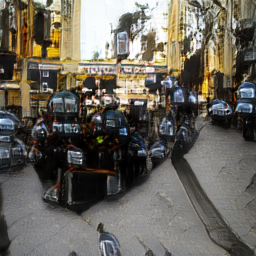}}
	\end{minipage} 
	&
	\begin{minipage}[b]{0.3\columnwidth}
		\centering
		\raisebox{-.5\height}{\includegraphics[width=\linewidth]{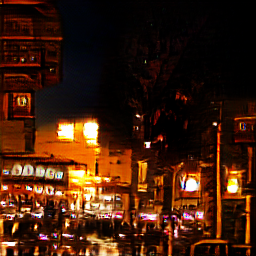}}
	\end{minipage} \\
  \hline
  
  \textbf{Many sheep are grazing for food in the grass.} 
    & 
    \begin{minipage}[b]{0.3\columnwidth}
		\centering
		\raisebox{-.5\height}{\includegraphics[width=\linewidth]{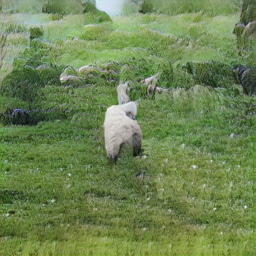}}
	\end{minipage} 
	&
	\begin{minipage}[b]{0.3\columnwidth}
		\centering
		\raisebox{-.5\height}{\includegraphics[width=\linewidth]{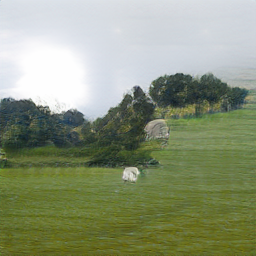}}
	\end{minipage} 
	&
	\begin{minipage}[b]{0.3\columnwidth}
		\centering
		\raisebox{-.5\height}{\includegraphics[width=\linewidth]{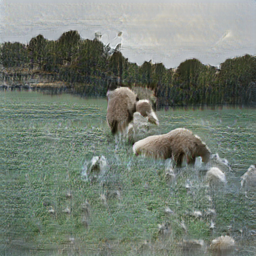}}
	\end{minipage} 
	&
	\begin{minipage}[b]{0.3\columnwidth}
		\centering
		\raisebox{-.5\height}{\includegraphics[width=\linewidth]{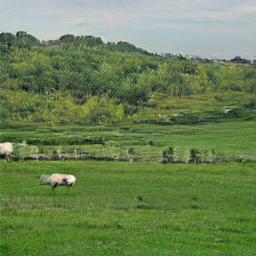}}
	\end{minipage} \\
  \hline
\end{tabular}
\end{center}
\end{table*}

\begin{table*}[t]
\centering
\begin{center}
\setlength{\abovecaptionskip}{0pt}%
\setlength{\belowcaptionskip}{10pt}%
\caption{Adjust the structural sequence of the Gradual refinement structure and compare the performance of the model under different network sequences.(I indicates Image Initialize stage. S indicates Sentence-level Refinement stage. W indicates Word-level Refinement stage.)} \label{tab:ablation}
\renewcommand{\arraystretch}{1.5}
\begin{tabular}{cccccccccc}
  \hline
  \hline
  & \multicolumn{3}{c}{Layer0} & \multicolumn{3}{c}{Layer1} & \multicolumn{3}{c}{Layer2}\\
  \hline
  & \textbf{FID} & \textbf{ITDis} & \textbf{CMD} & \textbf{FID} & \textbf{ITDis} & \textbf{CMD} & \textbf{FID} & \textbf{ITDis} & \textbf{CMD} \\
  \hline
  I$\rightarrow$S$\rightarrow$W & 169.13 & 2.05 & 36.71 & 61.37 & 1.15 & 15.67 & 22.58 & 0.80 & 8.04 \\
  \hline
  I$\rightarrow$W$\rightarrow$S & 168.52 & 2.04 & 36.66 & 63.54 & 1.12 & 16.62 & 25.93 & 0.85 & 9.13 \\
  \hline
  S$\rightarrow$W$\rightarrow$I & 182.12 & 2.18 & 36.98 & 69.33 & 1.18 & 16.81 & 34.99 & 0.92 & 10.31 \\
  \hline
\end{tabular}
\renewcommand{\arraystretch}{1.5}
\begin{tabular}{ccccc}
  \hline
  \textbf{Caption} & \textbf{Model} & \textbf{Layer0} & \textbf{Layer1} & \textbf{Layer2}\\
  \hline
  \multirow[m]{3}{0.12\linewidth}[-2cm]{\textbf{a man on a surf board rides a wave.}}
     & I$\rightarrow$S$\rightarrow$W & 
     \begin{minipage}[b]{0.3\columnwidth}
		\centering
		\raisebox{-.5\height}{\includegraphics[width=\linewidth]{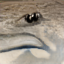}}
	\end{minipage}
     & 
     \begin{minipage}[b]{0.3\columnwidth}
		\centering
		\raisebox{-.5\height}{\includegraphics[width=\linewidth]{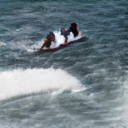}}
	\end{minipage}
	&
	\begin{minipage}[b]{0.3\columnwidth}
		\centering
		\raisebox{-.5\height}{\includegraphics[width=\linewidth]{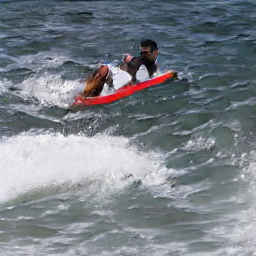}}
	\end{minipage}\\ 
	
     \cline{2-5}
     & I$\rightarrow$W$\rightarrow$S & 
     \begin{minipage}[b]{0.3\columnwidth}
		\centering
		\raisebox{-.5\height}{\includegraphics[width=\linewidth]{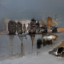}}
	\end{minipage}
     & 
     \begin{minipage}[b]{0.3\columnwidth}
		\centering
		\raisebox{-.5\height}{\includegraphics[width=\linewidth]{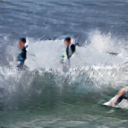}}
	\end{minipage}
	&
	\begin{minipage}[b]{0.3\columnwidth}
		\centering
		\raisebox{-.5\height}{\includegraphics[width=\linewidth]{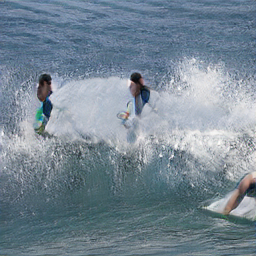}}
	\end{minipage}\\
	
     \cline{2-5}
     & S$\rightarrow$W$\rightarrow$I & 
     \begin{minipage}[b]{0.3\columnwidth}
		\centering
		\raisebox{-.5\height}{\includegraphics[width=\linewidth]{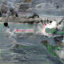}}
	\end{minipage}
     & 
     \begin{minipage}[b]{0.3\columnwidth}
		\centering
		\raisebox{-.5\height}{\includegraphics[width=\linewidth]{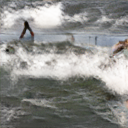}}
	\end{minipage}
	&
	\begin{minipage}[b]{0.3\columnwidth}
		\centering
		\raisebox{-.5\height}{\includegraphics[width=\linewidth]{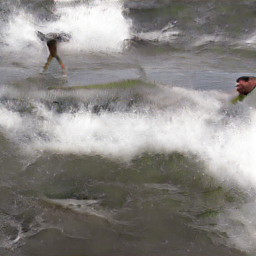}}
	\end{minipage}\\
  \hline
  
  \multirow[m]{3}{0.12\linewidth}[-2cm]{\textbf{A scenic view of a grassland with mountains in the background.}}
     & I$\rightarrow$S$\rightarrow$W & 
     \begin{minipage}[b]{0.3\columnwidth}
		\centering
		\raisebox{-.5\height}{\includegraphics[width=\linewidth]{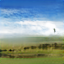}}
	\end{minipage}
     & 
     \begin{minipage}[b]{0.3\columnwidth}
		\centering
		\raisebox{-.5\height}{\includegraphics[width=\linewidth]{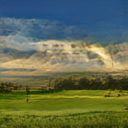}}
	\end{minipage}
	&
	\begin{minipage}[b]{0.3\columnwidth}
		\centering
		\raisebox{-.5\height}{\includegraphics[width=\linewidth]{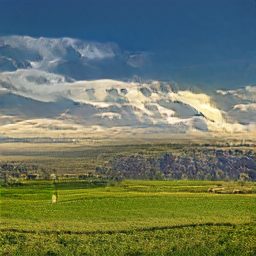}}
	\end{minipage}\\ 
	
     \cline{2-5}
     & I$\rightarrow$W$\rightarrow$S & 
     \begin{minipage}[b]{0.3\columnwidth}
		\centering
		\raisebox{-.5\height}{\includegraphics[width=\linewidth]{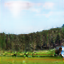}}
	\end{minipage}
     & 
     \begin{minipage}[b]{0.3\columnwidth}
		\centering
		\raisebox{-.5\height}{\includegraphics[width=\linewidth]{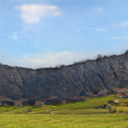}}
	\end{minipage}
	&
	\begin{minipage}[b]{0.3\columnwidth}
		\centering
		\raisebox{-.5\height}{\includegraphics[width=\linewidth]{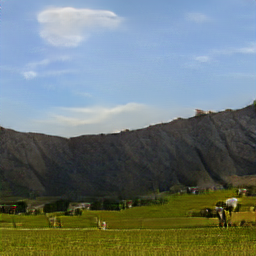}}
	\end{minipage}\\
	
     \cline{2-5}
     & S$\rightarrow$W$\rightarrow$I & 
     \begin{minipage}[b]{0.3\columnwidth}
		\centering
		\raisebox{-.5\height}{\includegraphics[width=\linewidth]{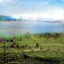}}
	\end{minipage}
     & 
     \begin{minipage}[b]{0.3\columnwidth}
		\centering
		\raisebox{-.5\height}{\includegraphics[width=\linewidth]{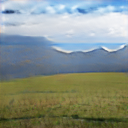}}
	\end{minipage}
	&
	\begin{minipage}[b]{0.3\columnwidth}
		\centering
		\raisebox{-.5\height}{\includegraphics[width=\linewidth]{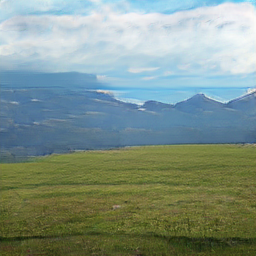}}
	\end{minipage}\\
	\hline
\end{tabular}
\end{center}
\end{table*}

\end{document}